\documentclass[runningheads]{llncs}

\usepackage{eccv}

\usepackage{eccvabbrv}

\usepackage{graphicx}
\usepackage{booktabs}

\usepackage{booktabs}
\usepackage{multirow}
\usepackage{array}

\usepackage{pifont}
\newcommand{\cmark}{\ding{51}}
\newcommand{\xmark}{\ding{55}}
\newcommand{\rc}[2]{#1/#2}

\usepackage[accsupp]{axessibility}  

\usepackage{hyperref}

\usepackage{orcidlink}

\begin{document}

\title{PriorPose: Reference-Guided Joint Deformation and Alignment for Category-Level Object Pose Estimation} 

\titlerunning{PriorPose}

\author{Yihan Chen\inst{1} \and
Huan Ren\inst{1} \and
Wenfei Yang \thanks{Corresponding author:~yangwf@ustc.edu.cn}\inst{1} \and
Hang Du\inst{2} \and
Tianzhu Zhang\inst{1} \and
Feng Wu\inst{1}}

\authorrunning{Y.~Chen et al.}

\institute{School of Information Science and Technology / National Key Laboratory of Deep Space Exploration, University of Science and Technology of China \and
Beijing Institute of Control Engineering\\
}

\maketitle
\begin{abstract}
Category-level object pose estimation seeks to recover a similarity transform $(R,t,s)$ for unseen instances without instance-specific CAD models. Most competitive methods are correspondence-based: \emph{prior-free} variants regress canonical (NOCS) coordinates directly from local observations and implicitly memorize the canonical frame in the weights, which ties the parameters to category-typical orientations and hurts generalization under distribution shift; \emph{prior-based} variants introduce a category prior but typically follow a serial deform-then-align pipeline, where underconstrained canonical completion can corrupt correspondences and induce error cascades in pose. We propose \textsc{PriorPose}, a reference-guided correspondence framework that keeps the category prior explicit and solves canonicalization and alignment jointly in a shared feature space. A reference-guided seeded transformer embeds the partial observation and the category prior as token sets and fuses them via geometry-aware seeds, from which the network jointly predicts a per-point NOCS field for visible points and a canonical deformation of the prior that reconstructs a full canonical instance, while a deep pose head regresses $(R,t,s)$ from the induced correspondences. A two-part \emph{shape consistency} objective, with canonical-space and camera-space consistency losses, couples correspondence, deformation, and pose, reducing reliance on memorized canonical orientations and avoiding deform-then-align error cascades. 
Experiments on standard and larger-category benchmarks demonstrate that PriorPose sets new state-of-the-art results on most evaluated metrics, especially under strict pose thresholds, while remaining competitive on relaxed pose and IoU metrics and showing improved robustness under shape variation and domain shift.
\end{abstract}

\section{Introduction}
\label{intro}

\begin{figure}[t]
  \centering
  \includegraphics[width=0.95\linewidth]{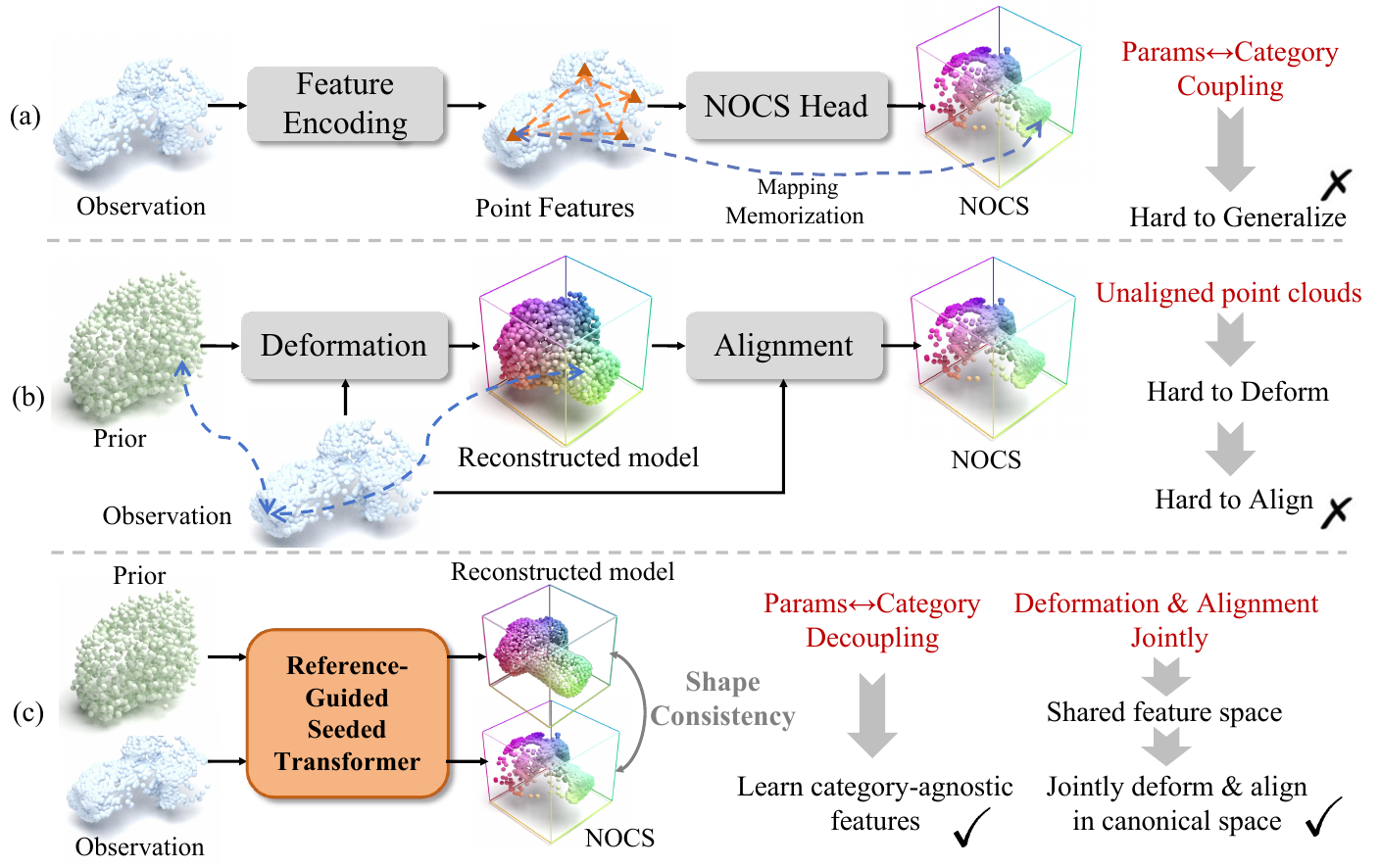} 
  \caption{Comparison of category-level correspondence-based methods.
(a) \emph{Prior-free} methods regress NOCS directly from local observations, which internalizes the canonical mapping in the weights and couples the parameters with category cues, hurting generalization.
(b) \emph{Prior-based} methods first deform a canonical instance model and then align by correspondences; unposed partial views make deformation ambiguous, coarse matches weaken alignment, and errors cascade.
(c) \textbf{Our method} embeds the prior and the observation in a shared space, performs joint deformation and alignment in the canonical space, enforces shape consistency, and learns category-agnostic weights, yielding stable correspondences under large shape variation and distribution shift.
}
  \label{fig:motivation}
\end{figure}

Category-level object pose estimation seeks to recover a similarity transform \((R,t,s)\)—rotation, translation, and metric size—for previously unseen instances within a category, \emph{without} requiring instance-specific CAD models~\cite{wang2019nocs,lin2022dpdn,lin2024agpose}. Compared with instance-level settings, it emphasizes intra-class generalization under unknown shape and scale, making it well suited to robotic manipulation, AR/VR, and mobile perception.

Prior work can be broadly categorized into two lines of approach:
(1) \emph{direct regression}, which learns a direct mapping from RGB-D or point cloud data to the pose 
\((R,t,s)\) in an end-to-end manner~\cite{dualposenet,fsnet,gpvpose,hspose,genpose,vinet,secondpose}, yet often struggles under challenging conditions such as partial visibility, scene clutter, and object symmetries.
(2) \emph{correspondence-based} methods, which first establish dense correspondences by predicting per-point coordinates in a category-level canonical space (e.g., NOCS)~\cite{wang2019nocs,liu2023istnet,lin2024agpose,ren2025spherepose,ren2025spotpose,ren2026compose}, and subsequently recover the object pose from these predicted correspondences.
The latter typically achieves higher accuracy and robustness when correspondences are reliable.

Despite the success of existing correspondence-based methods, there are two factors that limit the performance. 
\textbf{First}, \emph{prior-free} methods learn to regress canonical coordinates directly from local observed features, without an explicit shape reference, as shown in Figure~\ref{fig:motivation}(a)~\cite{lin2024agpose,ren2025spotpose}. This approach requires the model weights to memorize the mapping from local features to NOCS. A key challenge arises because the local features of the same part across different instances are highly similar while their NOCS coordinates differ. Consequently, the regression tends to bias towards category-typical averages, impairing generalization under significant shape variations. Moreover, since the model weights become tightly coupled with category-specific features, training independent NOCS prediction heads for each category restricts the model from leveraging cross-category data for improved generalization.
\textbf{Second}, \emph{prior-based} methods introduce a shape prior that helps decouple the weights from single-category cues and enables weight sharing~\cite{chen2020cass,tian2020spd,chen2021sgpa,lin2022dpdn}. They usually first produce a canonical instance in the category’s canonical space and then estimate pose from point-to-canonical correspondences (NOCS), as shown in Figure~\ref{fig:motivation}(b). However, both stages are fundamentally correspondence problems: deformation needs prior-to-instance correspondence, and pose recovery relies on observation-to-canonical correspondence. 
Consequently, it is ill-posed to split deformation and correspondence estimation into two different stages.  
Besides, deformation errors can directly bias the correspondences derived from the deformed template and then harm pose estimation performance.

To address the above challenges, our key idea is to use a category prior point cloud as guidance and to extract and fuse observation--prior features with a transformer-based network~\cite{vaswani2017attention}, while jointly optimizing deformation and alignment rather than fixing their order. This directly targets the two issues above.
\textbf{First}, introducing a category prior as an explicit canonical reference and fusing it with the partial observation gives the network a concrete ``standard'' orientation for each category, even when the current instance is only partially visible or geometrically different. Local features are interpreted relative to this shared canonical frame, rather than relying on implicit pose statistics stored in the weights, which reduces canonical pose memorization and makes it easier to share parameters across categories.
\textbf{Second}, from the same fused representation we jointly predict two complementary outputs in the canonical space: a per-point NOCS field for the visible observations and a canonical deformation of the prior that reconstructs a full instance. 
During training, the NOCS head uses the prior as a pose-aware reference so that local features on similar parts (e.g., mug handles) are encouraged to align to consistent canonical locations, while the deformation head, in turn, learns how the prior should bend toward the instance as these parts are progressively aligned in the canonical frame.
Since both predictions describe the same underlying geometry, we add lightweight shape-consistency terms that keep the NOCS field, the deformed prior, and the pose-aligned reconstruction in agreement with each other and with the observed points.
This joint, mutually guided prediction keeps correspondences and the completed shape aligned over the course of training, indirectly regularizes the learned pose, and reduces error accumulation compared with serial \emph{deform-then-align} pipelines that commit to an intermediate reconstruction before estimating pose.

To this end, we propose \textsc{PriorPose}, a single-stage model with \emph{category-agnostic} weights that keeps the canonical context external to the parameters. It solves canonicalization and alignment jointly in a shared feature space, as shown in Fig.~\ref{fig:motivation}(c).
\textbf{First}, a \textbf{reference-guided seeded transformer} embeds the observation and the category prior as token sets and lets them interact through a compact seed pool. Seed self-attention preserves within-stream structure, while token-to-seed attention enables cross-stream exchange. Within-stream kNN gating imposes a locality prior during token interaction. Geometry-quality gating computes a parameter-free score from same-stream seed neighborhoods to suppress unreliable tokens and stabilize attention routing.
\textbf{Second}, from the fused representation the network predicts two complementary outputs in a single forward pass: a per-point canonical correspondence field (NOCS) and a canonical deformation of the prior that reconstructs a full canonical instance.
A \textbf{shape-consistency} objective couples these outputs via \textbf{canonical-space consistency}, which ties the correspondence field to the reconstructed canonical instance, and \textbf{camera-space consistency}, which requires the reconstructed canonical shape, placed by the predicted pose $(R,t,s)$, to agree with the observed geometry.
These designs reduce memorization of category-specific orientations, avoid \emph{deform–then–align} error cascades, and yield more stable correspondences and poses under large shape variation and distribution shift.

The main contributions of this work can be summarized as follows:
(1) We analyze correspondence-based category-level pose estimation and identify two failure modes: \emph{prior-free} models internalize the canonical frame in the weights and overfit to category-typical orientations and shapes, while \emph{prior-based} serial deform--then--align pipelines treat deformation and alignment as separate steps of the same correspondence problem, leading to circular dependence and error cascades under partial, unaligned observations.
(2) We propose \textsc{PriorPose}, a reference-guided, single-stage model with \emph{category-agnostic} weights. It jointly predicts a per-point canonical correspondence field and a canonical deformation of the prior, coupled by shape-consistency terms in canonical and camera space. The seeded transformer uses within-stream kNN gating and geometry-quality gating to stabilize attention routing under sparse or corrupted depth.
(3) Extensive experiments on standard, cross-dataset, and larger-category benchmarks demonstrate state-of-the-art results on most metrics, consistent gains under strict pose criteria, and the effectiveness of the proposed joint deformation-alignment formulation.

\section{Related Work}

\paragraph{Prior-free Correspondence Methods.}
NOCS~\cite{wang2019nocs} popularized learning a category-shared canonical space and regressing dense correspondences for pose and size, and subsequent prior-free methods improve correspondence quality without explicit templates.
IST-Net~\cite{liu2023istnet} learns an implicit space transformation into the canonical space.
AG-Pose~\cite{lin2024agpose} detects instance-adaptive keypoints and aggregates local-to-global geometry.
SpherePose~\cite{ren2025spherepose} introduces shared proxies with rotation-aware features to reduce shape dependence.
SpotPose~\cite{ren2025spotpose} revisits correspondence with outlier suppression and robust fitting.
These approaches are accurate and efficient, but because the canonical frame is internalized in the weights, predictions can drift toward training-set canonical modes under large shape variation or distribution shift. 

\paragraph{Prior-based and Shape-Prior Methods.}
Another line leverages class priors such as mean shapes or prototypes as explicit canonical references to inject global semantics and complete missing geometry before alignment.
Early deform--then--align pipelines include CASS~\cite{chen2020cass}, SPD~\cite{tian2020spd}, SGPA~\cite{chen2021sgpa}, and DPDN~\cite{lin2022dpdn}.
More recently, GCE-Pose~\cite{li2025gcepose} reconstructs global semantics and geometry from category prototypes and fuses them with partial observations to improve correspondence and pose recovery.
However, single-view canonical reconstruction is highly underconstrained under partial views when the canonical alignment is unknown.
Errors in the reconstructed model can bias correspondences and propagate to downstream pose estimation, making performance sensitive to prior quality and deformation accuracy.
This serial separation can amplify errors and lead to cascades when observations are partial and unaligned.

\paragraph{Joint shape and pose estimation.}
Several works jointly estimate category-level shape and pose using object-centric latent or implicit representations, often in multi-object settings, and differ from NOCS-style correspondence learning in problem setup, output representation, and supervision.
Examples include CenterSnap~\cite{irshad2022centersnap}, ShAPO~\cite{irshad2022shapo}, CARTO~\cite{heppert2023carto}, and FSD~\cite{lunayach2024fsd}.
Related lines often regress canonical depth or shape representations and use them to facilitate pose recovery, e.g., ACR-Pose~\cite{fan2024acr,fan2022object}, while SSP-Pose~\cite{zhang2022ssp} designs symmetry-aware shape-prior deformation losses to mitigate pose ambiguity.
Our work instead stays in the correspondence regime and uses an explicit mean-shape reference for token-level observation–prior interaction, with lightweight shape-consistency constraints in canonical and camera spaces.

\paragraph{Direct Regression.}
Direct regression methods predict category-level 6-DoF pose and metric size from RGB-D or point clouds without explicit correspondences, e.g., DualPoseNet~\cite{dualposenet}, FS-Net~\cite{fsnet}, GPV-Pose~\cite{gpvpose}, HS-Pose~\cite{hspose}, GenPose~\cite{genpose}, VI-Net~\cite{vinet}, and SecondPose~\cite{secondpose}.
They are efficient but can degrade under partial views, heavy clutter, and strong symmetries.
Our work instead targets the correspondence family and focuses on making the canonical reference explicit while jointly coupling partial NOCS with canonical deformation and pose.

\begin{figure}[t]
  \centering
  \includegraphics[width=1.0\linewidth]{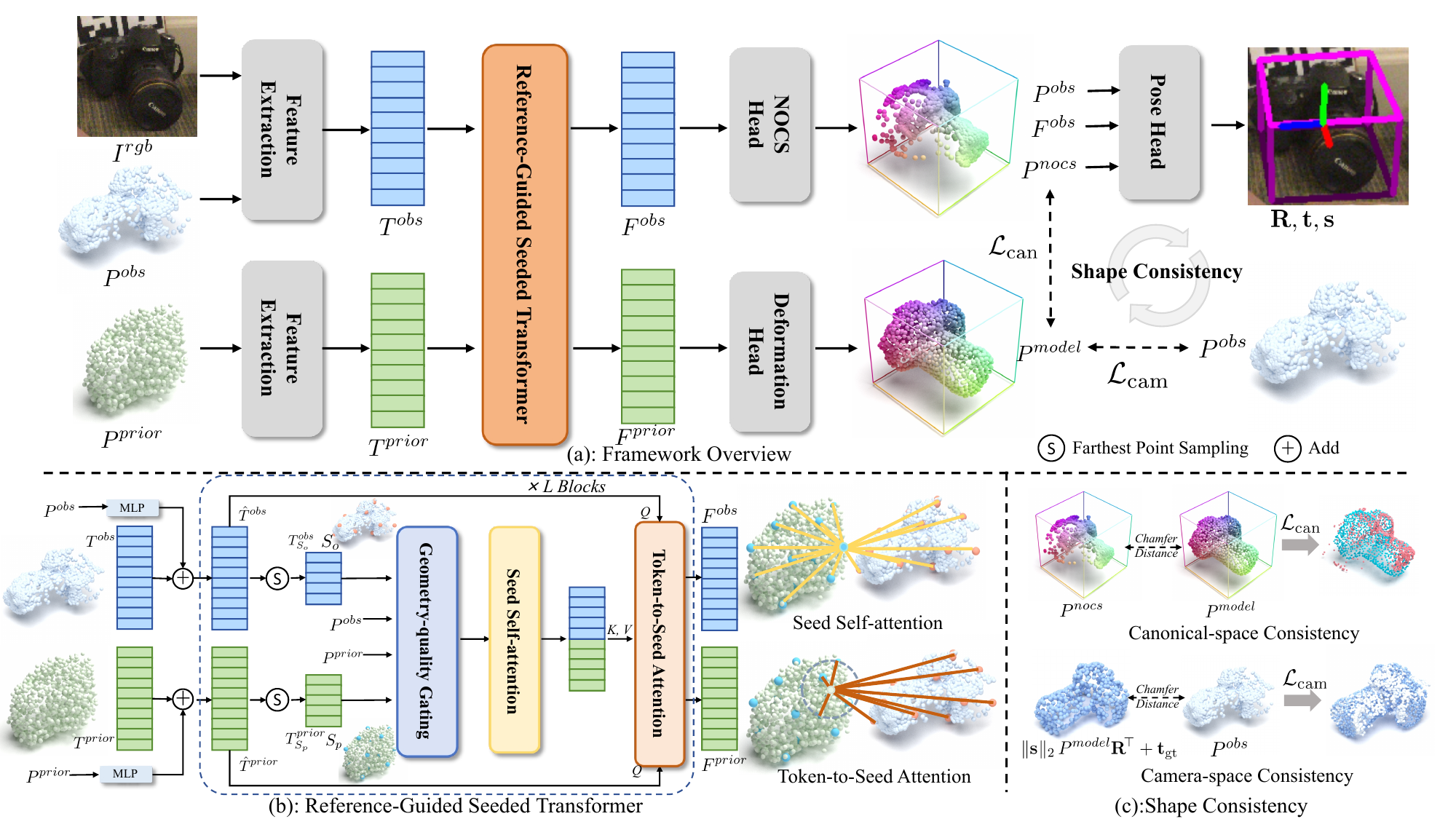} 
\caption{
\textbf{Overview of \textsc{PriorPose}.}
(a) We use a category prior, the mean shape, as an explicit canonical reference: features from the partial observation and the category prior are fused in a shared space by a reference-guided seeded transformer, which jointly predicts per-point NOCS, a canonical deformation, and the pose $(R,t,s)$, trained with shape consistency.
(b) Details of the reference-guided seeded transformer with geometry-aware seeds and kNN-gated attention for efficient observation–prior interaction.
(c) Shape consistency, enforced by canonical-space and camera-space losses that jointly couple NOCS, deformation and pose.
}

  \label{fig:fft}
\end{figure}

\section{Method}
\noindent\textbf{Task Definition.}
Given an RGB-D image, we first obtain instance masks with an off-the-shelf instance segmentation network~\cite{he2017mask} and crop the RGB to $I^{\mathrm{rgb}}\!\in\!\mathbb{R}^{H\times W\times 3}$.
Back-projecting the segmented depth yields an observed point cloud $P^{\mathrm{obs}}\!\in\!\mathbb{R}^{N_{\mathrm{obs}}\times 3}$.
With a category prior point set $P^{\mathrm{prior}}\!\in\!\mathbb{R}^{N_{\mathrm{prior}}\times 3}$ as canonical context, the goal is to estimate rotation $R\!\in\!\mathrm{SO}(3)$, translation $t\!\in\!\mathbb{R}^3$, and size $s\!\in\!\mathbb{R}^3$ of the instance.

\noindent\textbf{Overview.}
As illustrated in Figure~\ref{fig:fft}, our framework has three components:
(1) \emph{Feature extraction} from the cropped RGB image $I^{\mathrm{rgb}}$, the observation point cloud $P^{\mathrm{obs}}$, and the category prior $P^{\mathrm{prior}}$ (Sec.~\ref{sec:feat});
(2) \emph{Reference-guided seeded fusion and prediction heads}, where a seeded transformer fuses observation and prior tokens with within-stream kNN gating and geometry-quality gating (Sec.~\ref{sec:fusion}).
The fused features are passed to a correspondence head that predicts per-point NOCS, a deformation head that reconstructs a canonical instance, and a deep pose head that regresses $(R,t,s)$ from the induced correspondences (Sec.~\ref{sec:head}).
(3) \emph{Shape-consistent training objectives} that couple correspondence, deformation, and pose via NOCS and pose losses, a canonical reconstruction loss, and two shape-consistency terms in canonical and camera space (Sec.~\ref{sec:losses}).

\subsection{Feature Extraction}
\label{sec:feat}

\noindent\textbf{Inputs.}
From a segmented RGB-D frame, we obtain a cropped RGB image $I^{\mathrm{rgb}} \in \mathbb{R}^{H\times W\times 3}$ and a partial object point cloud $P^{\mathrm{obs}}\!=\!\{p_i\}_{i=1}^{N_{\mathrm{obs}}}\!\in\!\mathbb{R}^{N_{\mathrm{obs}}\times 3}$.
For each category $c$, a canonical mean shape (prior) is represented by a point set $P^{\mathrm{prior}}\!=\!\{\bar{p}_j\}_{j=1}^{N_{\mathrm{prior}}}\!\in\!\mathbb{R}^{N_{\mathrm{prior}}\times 3}$.

\paragraph{Observation tokens.}
We extract pointwise geometric features with PointNet++~\cite{qi2017pointnet++},
$F^{\mathrm{geo}}=\mathrm{PN2}(P^{\mathrm{obs}})\in\mathbb{R}^{N_{\mathrm{obs}}\times D_g}$.
Semantic image features are produced by DINOv2~\cite{oquab2023dinov2} from $I^{\mathrm{rgb}}$ and associated with points via camera projection,
$F^{\mathrm{img}}\in\mathbb{R}^{N_{\mathrm{obs}}\times D_i}$.
We further encode 3D coordinates with an MLP,
$F^{\mathrm{pos}}=\mathrm{MLP}^{\mathrm{obs}}_{\text{pos}}(P^{\mathrm{obs}})\in\mathbb{R}^{N_{\mathrm{obs}}\times D_p}$.
After channel-wise concatenation and a linear projection, we form observation tokens:
\begin{equation}
\small
T^{\mathrm{obs}}=\mathrm{Proj}\!\big([F^{\mathrm{geo}}\|\!F^{\mathrm{img}}\|\!F^{\mathrm{pos}}]\big)\in\mathbb{R}^{N_{\mathrm{obs}}\times D}.
\label{eq:obs_token}
\end{equation}

\paragraph{Prior tokens.}
For the category prior, we allocate a learnable token per prior point,
$E^{\mathrm{learn}}\in\mathbb{R}^{N_{\mathrm{prior}}\times D}$,
and add a geometric embedding of the prior coordinates,
$E^{\mathrm{geo}}=\mathrm{MLP}^{\mathrm{prior}}_{\text{pos}}(P^{\mathrm{prior}})\in\mathbb{R}^{N_{\mathrm{prior}}\times D}$.
The prior tokens are
\begin{equation}
\small
T^{\mathrm{prior}}=\mathrm{Proj}\!\big([E^{\mathrm{learn}}\|\!E^{\mathrm{geo}}]\big)\in\mathbb{R}^{N_{\mathrm{prior}}\times D}.
\label{eq:prior_token}
\end{equation}

\paragraph{Token set.}
We use observation tokens $T^{\mathrm{obs}}\in\mathbb{R}^{N_{\mathrm{obs}}\times D}$ and prior tokens $T^{\mathrm{prior}}\in\mathbb{R}^{N_{\mathrm{prior}}\times D}$.
We add learnable source tags $e^{\mathrm{obs}},e^{\mathrm{prior}}\in\mathbb{R}^{D}$ to mark stream identity:
$\tilde{T}^{\mathrm{obs}}=T^{\mathrm{obs}}+(e^{\mathrm{obs}})^{\top}$ and
$\tilde{T}^{\mathrm{prior}}=T^{\mathrm{prior}}+(e^{\mathrm{prior}})^{\top}$.
The fused token sequence is $T=[\tilde{T}^{\mathrm{obs}};\tilde{T}^{\mathrm{prior}}]\in\mathbb{R}^{L\times D}$ with $L=N_{\mathrm{obs}}+N_{\mathrm{prior}}$.

\subsection{Reference-Guided Seeded Transformer}
\label{sec:fusion}
\vspace{-3pt}
\noindent\textbf{Overview.}
Starting from $T=[\tilde T^{\mathrm{obs}};\tilde T^{\mathrm{prior}}]\in\mathbb{R}^{L\times D}$ with source tags, we sample $K$ seed indices once by FPS on $P^{\mathrm{obs}}$ and $P^{\mathrm{prior}}$, gather the corresponding tokens to form $Z\in\mathbb{R}^{K\times D}$, and reuse the seed indices across all blocks~\cite{zhou2022seedformer}.
Each block alternates (i) seed self-attention to update $Z$ with global context, and (ii) token-to-seed attention that routes messages from all tokens to seeds.
In routing, we apply within-stream $k$NN gating to enforce geometric locality, while keeping cross-stream seeds visible, and add a geometry-quality gating as a logit bias to down-weight unreliable associations.
This seeded design yields efficient observation--prior interaction with $O(LK)$ attention cost where $K\!\ll\!L$.

\vspace{-5pt}
\paragraph{Seeded attention with an additive mask.}
With \(H\) heads and per-head dimension \(d=D/H\), we use masked dot-product attention:
\begin{equation}
\label{eq:attn}
\ Attn(Q,K,V;M)=\ softmax ~\!\Big(\tfrac{QK^\top}{\sqrt{d}}+M\Big)V.
\end{equation}
The additive mask \(M\) biases attention logits and uses \(-\infty\) to forbid connections.

\vspace{-5pt}
\paragraph{Geometry-aware seed selection.}
We choose \(K_o\) observation seeds and \(K_p\) prior seeds by FPS on coordinates:
\vspace{-4pt}
\[
S_o=\mathrm{FPS}\!\big(P^{\mathrm{obs}},K_o\big),\qquad
S_p=\mathrm{FPS}\!\big(P^{\mathrm{prior}},K_p\big).
\vspace{-4pt}
\]
We initialize the seed pool from tagged tokens:
\[
Z=[\,\tilde T^{\mathrm{obs}}_{S_o};\,\tilde T^{\mathrm{prior}}_{S_p}\,]\in\mathbb{R}^{K\times D},
\qquad K=K_o+K_p.
\]
Seed indices are fixed across blocks and only seed features are updated.

\vspace{-5pt}
\paragraph{Within-stream \(k\)NN gating and cross-stream visibility.}
We apply \(k\)NN gating only within each stream to encode a geometric locality prior.
Cross-stream Euclidean distances are not meaningful before pose is estimated.
We therefore gate only same-source token--seed pairs and keep all cross-source seeds visible.

We view the seed pool as $Z=[Z^{\mathrm{obs}};Z^{\mathrm{prior}}]$ with $K=K_o+K_p$.
For an observation token $i$, let $\mathcal{N}_k(i;S_o)\subseteq\{1,\ldots,K_o\}$ be its $k$ nearest observation seeds in $P^{\mathrm{obs}}$, and allow attention to these seeds and all prior seeds:
\[
\mathcal{A}_i=\mathcal{N}_k(i;S_o)\ \cup\ \{K_o+1,\ldots,K\}.
\]
For a prior token $j$, we define $\mathcal{A}_{N_{\mathrm{obs}}+j}$ symmetrically using $k$NN prior seeds in $P^{\mathrm{prior}}$ and keep all observation seeds visible.
We implement hard gating by an additive mask
\[
M^{\mathrm{knn}}(t,u)=
\begin{cases}
0,& u\in\mathcal{A}_t,\\
-\infty,& \text{otherwise}.
\end{cases}
\]

\vspace{-5pt}
\paragraph{Geometry-quality gating.}
We further suppress unreliable tokens by a parameter-free geometry-quality score computed from same-stream seed neighborhoods.
For conciseness we describe the observation stream, and apply the same procedure to the prior stream.
We define $\hat q_j$ for each prior point $\bar p_j$ analogously using $k$NN prior seeds in $P^{\mathrm{prior}}$.

For an observation point \(p_i\in P^{\mathrm{obs}}\), let \(\mathcal{N}_k(i;S_o)\) be its \(k\) nearest observation seeds.
Let $\mathbf{s}_u \in \mathbb{R}^3$ denote the 3D coordinate of the $u$-th observation seed in $P^{\mathrm{obs}}_{S_o}$. 
Here $u$ indexes the observation seeds selected by $S_o$, and $\mathcal{N}_k(i;S_o)$ returns indices in this seed set.
We define the average seed distance
\[
d_i=\frac{1}{k}\sum_{u\in\mathcal{N}_k(i;S_o)}\|p_i-\mathbf{s}_u\|_2,
\qquad
\sigma=\frac{1}{N_{\mathrm{obs}}}\sum_i d_i,
\qquad
q_i=\exp\!\Big(-\tfrac{1}{2}(d_i/\sigma)^2\Big),
\]
and normalize it to \([0,1]\):
\[
\hat q_i=\mathrm{clip}\!\Big(\frac{q_i}{\frac{1}{N_{\mathrm{obs}}}\sum_i q_i+\epsilon},\,0,\,1\Big).
\]
We use \(\hat q\) in two places.
First, we gate tokens before token-to-seed attention:
\[
\tilde T^{\mathrm{obs}}_i \leftarrow \hat q_i\,\tilde T^{\mathrm{obs}}_i,
\qquad
\tilde T^{\mathrm{prior}}_j \leftarrow \hat q_j\,\tilde T^{\mathrm{prior}}_j.
\]
Second, we add a geometry-quality bias on attention logits:
\begin{equation}\label{eq:score_bias}
M^{\mathrm{geo}}(t,u)=\gamma\big(\hat q_t+\hat q_u\big),
\qquad
\hat q_t,\hat q_u\in[0,1].
\end{equation}
Here \(\gamma>0\) controls the strength.
For a seed key $u$, we set $\hat q_u$ to the geometry-quality score of the corresponding seed point (i.e., $\hat q$ evaluated at that seed) in the same stream.
The total mask in Eq.~\eqref{eq:attn} is \(M=M^{\mathrm{knn}}+M^{\mathrm{geo}}\).

\paragraph{Block updates.}
Each block updates seeds by self-attention and then updates all tokens by token-to-seed attention using the total mask $M$ in Eq.~\eqref{eq:attn}.
We use standard pre-norm Transformer blocks with residual MLPs.

\subsection{Prediction Heads}
\label{sec:head}
\paragraph{NOCS and deformation heads.}
After seeded fusion we obtain updated features for the two streams.
Let $F^{\mathrm{obs}}=\{F^{\mathrm{obs}}_i\}_{i=1}^{N_{\mathrm{obs}}}$ be the observation features and $F^{\mathrm{prior}}=\{F^{\mathrm{prior}}_j\}_{j=1}^{N_{\mathrm{prior}}}$ be the prior features.
The NOCS head predicts per-observation NOCS:
\[P^{\mathrm{nocs}}=\mathrm{MLP}_{\mathrm{nocs}}(F^{\mathrm{obs}})\in\mathbb{R}^{N_{\mathrm{obs}}\times 3}.
\vspace{-6pt}\]
The deformation head predicts displacements for prior points and forms a canonical instance:
\[\hat{\Delta}=\mathrm{MLP}_{\mathrm{def}}(F^{\mathrm{prior}})\in\mathbb{R}^{N_{\mathrm{prior}}\times 3},
P^\mathrm{model}=P^{\mathrm{prior}}+\hat{\Delta}.
\vspace{-5pt}\]
Here $P^{\mathrm{nocs}}$ are the NOCS coordinates and $P^\mathrm{model}$ is the reconstructed canonical instance.

\paragraph{Deep pose regressor.}
Given observation points $P^{\mathrm{obs}}$ with fused features $F^{\mathrm{obs}}$ and their predicted NOCS $P^{\mathrm{nocs}}$, we estimate pose and size following previous work~\cite{lin2024agpose,lin2022dpdn} using a deep estimator.
Three MLP heads predict rotation, translation, and size:
\begin{equation}\label{eq:fpose}
f_{\mathrm{pose}}
= \operatorname{concat}\!\big[\,P^{\mathrm{nocs}},\,P^{\mathrm{obs}},\,F^{\mathrm{obs}}\big].
\end{equation}
\begin{equation}\label{eq:rtspred}
\mathbf{R},\, \mathbf{t},\, \mathbf{s}
= \mathrm{MLP}_{R}\!\big(f_{\mathrm{pose}}\big),\;
  \mathrm{MLP}_{t}\!\big(f_{\mathrm{pose}}\big),\;
  \mathrm{MLP}_{s}\!\big(f_{\mathrm{pose}}\big).
\end{equation}

\subsection{Shape-Consistent Training Objectives}
\label{sec:losses}
We use three standard loss components together with two shape-consistency terms.
First, we adopt a NOCS regression loss~\cite{wang2019nocs} and a symmetry-aware pose loss
(\emph{cf.}~\cite{yang2024ps6d}).
Second, we add a reconstruction loss on the canonical instance via the Chamfer distance
(\emph{cf.}~\cite{fan2017pointset,achlioptas2018learning}).
Third, we impose the shape consistency:
a \textbf{canonical-space consistency} term that ties the correspondence field to the reconstructed canonical instance,
and a \textbf{camera-space consistency} term that enforces agreement between the reconstructed canonical shape,
placed by the predicted pose, and the observed geometry.

\paragraph{NOCS regression.}
Let $P^{\mathrm{obs}}\in\mathbb{R}^{N_{\mathrm{obs}}\times 3}$ be the camera-space observations and
$P^{\mathrm{nocs}}\in\mathbb{R}^{N_{\mathrm{obs}}\times 3}$ the predicted NOCS.
With ground-truth $T_{\mathrm{gt}}=(\mathbf{R}_{\mathrm{gt}},\mathbf{t}_{\mathrm{gt}},\mathbf{s}_{\mathrm{gt}})$, the NOCS targets are obtained by the inverse transform:
\[
P^{\mathrm{nocs}}_{gt}
= \|\mathbf{s}_{\mathrm{gt}}\|_2^{-1}\big(P^{\mathrm{obs}}-\mathbf{1}\mathbf{t}_{\mathrm{gt}}^{\!\top}\big)\,\mathbf{R}_{\mathrm{gt}},
\]
where $\mathbf{1}$ denotes an all-ones column vector for broadcasting.
We supervise $P^{\mathrm{nocs}}$ with a point-wise Smooth-$L_1$ loss and reweight points to reduce the impact of noisy correspondences.
Let $e_i$ be the mean Smooth-$L_1$ error over the three NOCS coordinates for point $i$, and let $r_i\in[0,1]$ be a predicted per-point reliability score.
We compute stop-gradient, mean-normalized weights
\[
w_i=\frac{\mathrm{stopgrad}(r_i)}{\frac{1}{N_{\mathrm{obs}}}\sum_j \mathrm{stopgrad}(r_j)+\epsilon}.
\]
The weighted NOCS loss is
\begin{equation}
\mathcal{L}_{\mathrm{nocs}}
=\frac{\sum_i w_i\,e_i}{\sum_i w_i+\epsilon}.
\label{eq:nocs_w}
\end{equation}
This weighting follows prior robust correspondence learning with per-point reliability~\cite{ren2025spotpose, chen2025structure}.
We include a lightweight self-calibration for $r_i$ in the supplementary.

\paragraph{Symmetry-aware pose.}
Following \cite{yang2024ps6d}, we measure rotation under category symmetries and use $\ell_2$ terms for translation and size:
\begin{equation}
\mathcal{L}_{\mathrm{pose}}
= \lambda_R \min_{\mathbf{R}_s\in\mathcal{R}_s}\!\big\|\mathbf{R}_{\mathrm{gt}}\mathbf{R}_s-\mathbf{R}\big\|_{F}
+ \lambda_t \|\mathbf{t}-\mathbf{t}_{\mathrm{gt}}\|_{2}
+ \lambda_s \|\mathbf{s}-\mathbf{s}_{\mathrm{gt}}\|_{2}.
\end{equation}
Here $\mathbf{R}\!\in\!\mathrm{SO}(3)$, $\mathbf{t}$, $\mathbf{s}\!\in\!\mathbb{R}^3$ are the predicted rotation, translation, and size; $(\mathbf{R}_{\mathrm{gt}},\mathbf{t}_{\mathrm{gt}},\mathbf{s}_{\mathrm{gt}})$ are ground truth; $\mathcal{R}_s$ is the discrete symmetry set (identity if none); $\|\cdot\|_F$ and $\|\cdot\|_2$ denote Frobenius and Euclidean norms; $\lambda_R,\lambda_t,\lambda_s$ are weights.

\paragraph{Canonical reconstruction.}
We supervise the reconstructed canonical instance using the squared Chamfer distance~\cite{fan2017pointset,achlioptas2018learning}:
\begin{equation}
\mathcal{L}_{\mathrm{rec}}=\mathrm{CD}(P^{\mathrm{model}},P^{\mathrm{model}}_{\mathrm{gt}}).
\end{equation}

\newcommand{\cdone}[2]{\frac{1}{|#1|}\sum_{\mathbf{a}\in #1}\min_{\mathbf{b}\in #2}\|\mathbf{a}-\mathbf{b}\|_2^2}
\paragraph{Canonical-space consistency.}
Since $P^{\mathrm{nocs}}$ and $P^{\mathrm{model}}$ lie in the same canonical space, we enforce their agreement while accounting for the fact that $P^{\mathrm{nocs}}$ is partial and $P^{\mathrm{model}}$ is complete.
During training we remove outliers in the predicted correspondences by a nearest-neighbor distance threshold to the ground-truth canonical model~\cite{lin2024agpose}.
Specifically, we discard NOCS points whose nearest distance to $P^{\mathrm{model}}_{\mathrm{gt}}$ exceeds $\tau_{\mathrm{can}}$ and denote the remaining set as $P^{\mathrm{nocs}}_{\mathrm{flt}}$.
We then measure a one-sided Chamfer distance from the filtered NOCS points to the reconstructed canonical instance:
\begin{equation}
\mathcal{L}_{\mathrm{can}}
=\cdone{P^{\mathrm{nocs}}_{\mathrm{flt}}}{P^{\mathrm{model}}}.
\end{equation}

\paragraph{Camera-space consistency.}
We enforce consistency in the camera space between the reconstructed canonical instance placed by the predicted pose and the observed geometry.
We first place the ground-truth canonical model into the camera frame:
\[
\widetilde{P}^{\mathrm{model}}_{\mathrm{gt}}
=\|\mathbf{s}_{\mathrm{gt}}\|_2\,P^{\mathrm{model}}_{\mathrm{gt}}\mathbf{R}_{\mathrm{gt}}^{\!\top}
+\mathbf{1}\mathbf{t}_{\mathrm{gt}}^{\!\top}.
\]
At training time, we filter observations by retaining points whose nearest distance to $\widetilde{P}^{\mathrm{model}}_{\mathrm{gt}}$ is below $\tau_{\mathrm{cam}}$, denoted as $P^{\mathrm{obs}}_{\mathrm{flt}}$.
We then measure a one-sided Chamfer distance from the filtered observations to the reconstructed instance placed by the predicted pose:
\begin{equation}
\mathcal{L}_{\mathrm{cam}}
=\cdone{P^{\mathrm{obs}}_{\mathrm{flt}}}
{\|\mathbf{s}\|_2P^{\mathrm{model}}\mathbf{R}^{\!\top}+\mathbf{1}\mathbf{t}^{\!\top}}.
\end{equation}

\paragraph{Total loss.} The overall loss function is as follows:
\begin{equation}
\mathcal{L}
= \lambda_{\mathrm{1}}\mathcal{L}_{\mathrm{nocs}}
+ \lambda_{\mathrm{2}}\mathcal{L}_{\mathrm{pose}}
+ \lambda_{\mathrm{3}}\mathcal{L}_{\mathrm{rec}}
+ \lambda_{\mathrm{4}}\mathcal{L}_{\mathrm{can}}
+ \lambda_{\mathrm{5}}\mathcal{L}_{\mathrm{cam}}.
\end{equation}

\section{Experiments}

\subsection{Experimental Setup}
\label{sec:exp-setup}

\paragraph{Datasets and metrics.}
We evaluate on CAMERA25 and REAL275 from NOCS~\cite{wang2019nocs}, HouseCat6D~\cite{jung2024housecat6d}, Wild6D~\cite{fu2022category}, and two larger-category benchmarks, Omni6DPose~\cite{zhang2024omni6dpose} and PACE~\cite{you2024pace}.
CAMERA25/REAL275 are the standard synthetic-to-real NOCS benchmarks; HouseCat6D contains stronger occlusion and intra-class variation; Wild6D is used for zero-shot cross-dataset evaluation.
Following prior work~\cite{lin2024agpose,ren2025spotpose}, we report pose mAP under $(5^\circ,2\mathrm{cm})$, $(5^\circ,5\mathrm{cm})$, $(10^\circ,2\mathrm{cm})$, and $(10^\circ,5\mathrm{cm})$, and 3D IoU mAP at $50\%$ and $75\%$. Symmetric objects follow the standard symmetry-aware protocol.
For Omni6DPose and PACE, we follow their official category-level metrics.

\paragraph{Implementation details.}
We use the same instance masks as prior work~\cite{lin2024agpose,ren2025spotpose}.
Each observation contains $N_{\mathrm{obs}}{=}1024$ points and a $224{\times}224$ RGB crop; each category prior contains $N_{\mathrm{prior}}{=}512$ points.
Observation tokens concatenate PointNet++ features, DINOv2 image features, and coordinate embeddings into $D{=}256$ dimensions.
We use $6$ transformer blocks, $4$ heads, $K_o{=}96$, $K_p{=}64$, $k{=}32$, and $\gamma{=}0.2$.
Training details and loss weights are in the supplementary material.

\begin{table}[t]
\caption{Performance comparison on REAL275 and CAMERA25. Each entry is reported as REAL275/CAMERA25. Best is in \textbf{bold}, second-best is \underline{underlined}.}
\label{tab:main_results}
\centering
\scriptsize
\setlength{\tabcolsep}{1.8pt}
\renewcommand{\arraystretch}{0.98}

\begin{tabular}{@{} >{\raggedright\arraybackslash}p{2.5cm} c
                c c c c c c @{}}
\toprule
\textbf{Method} & \textbf{Prior} &
IoU$_{50}$ & IoU$_{75}$ &
$5^{\circ}$2cm & $5^{\circ}$5cm &
$10^{\circ}$2cm & $10^{\circ}$5cm \\
\midrule

\multicolumn{8}{@{}l}{\textit{Without prior}}\\
GenPose\cite{genpose} & \xmark &
\rc{--}{--} & \rc{--}{--} &
\rc{52.1}{79.9} & \rc{60.9}{\underline{84.4}} & \rc{72.4}{84.6} & \rc{84.0}{89.6} \\
VI-Net\cite{vinet} & \xmark &
\rc{--}{--} & \rc{--}{--} &
\rc{50.0}{74.1} & \rc{57.6}{81.4} & \rc{70.8}{79.3} & \rc{82.1}{87.3} \\
SecondPose\cite{secondpose} & \xmark &
\rc{--}{--} & \rc{--}{--} &
\rc{56.2}{--} & \rc{63.6}{--} & \rc{74.7}{--} & \rc{86.0}{--} \\
NOCS\cite{wang2019nocs} & \xmark &
\rc{78.0}{83.9} & \rc{30.1}{69.5} &
\rc{7.2}{32.3} & \rc{10.0}{40.9} & \rc{13.8}{48.2} & \rc{25.2}{64.4} \\
IST-Net\cite{liu2023istnet} & \xmark &
\rc{82.5}{93.7} & \rc{76.6}{90.8} &
\rc{47.5}{71.3} & \rc{53.4}{79.9} & \rc{72.1}{79.4} & \rc{80.5}{89.9} \\
Query6DoF\cite{wang2023query6dof} & \xmark &
\rc{82.5}{91.9} & \rc{76.1}{88.1} &
\rc{49.0}{78.0} & \rc{58.9}{83.1} & \rc{68.7}{83.9} & \rc{83.0}{90.0} \\
AG-Pose\cite{lin2024agpose} & \xmark &
\rc{\underline{83.7}}{93.8} & \rc{79.5}{91.3} &
\rc{54.7}{77.8} & \rc{61.7}{82.8} & \rc{74.7}{85.5} & \rc{83.1}{91.6} \\
SpherePose\cite{ren2025spherepose} & \xmark &
\rc{\textbf{84.1}}{\textbf{94.8}} & \rc{\textbf{81.2}}{\underline{92.4}} &
\rc{58.2}{78.3} & \rc{\underline{67.4}}{84.3} & \rc{76.2}{84.8} & \rc{\underline{88.2}}{\underline{92.3}} \\
SpotPose\cite{ren2025spotpose} & \xmark &
\rc{\textbf{84.1}}{\underline{94.3}} & \rc{\textbf{81.2}}{\textbf{92.5}} &
\rc{\underline{59.7}}{\underline{80.4}} & \rc{64.8}{83.8} & \rc{\textbf{81.5}}{\underline{87.7}} & \rc{\underline{88.2}}{92.2} \\
\addlinespace[1pt]

\multicolumn{8}{@{}l}{\textit{With prior}}\\
SPD\cite{tian2020spd} & \cmark &
\rc{77.3}{93.2} & \rc{53.2}{83.1} &
\rc{19.3}{54.3} & \rc{21.4}{59.0} & \rc{43.2}{73.3} & \rc{54.1}{81.5} \\
SGPA\cite{chen2021sgpa} & \cmark &
\rc{80.1}{93.2} & \rc{61.9}{88.1} &
\rc{35.9}{70.7} & \rc{39.6}{74.5} & \rc{61.3}{82.7} & \rc{70.7}{88.4} \\
SAR-Net\cite{lin2022sar} & \cmark &
\rc{79.3}{86.8} & \rc{62.4}{79.0} &
\rc{31.6}{66.7} & \rc{42.3}{70.9} & \rc{50.3}{75.3} & \rc{68.3}{80.3} \\
DPDN\cite{lin2022dpdn} & \cmark &
\rc{83.4}{--} & \rc{76.0}{--} &
\rc{46.0}{--} & \rc{50.7}{--} & \rc{70.4}{--} & \rc{78.4}{--} \\
GCE-Pose\cite{li2025gcepose} & \cmark &
\rc{\textbf{84.1}}{--} & \rc{\underline{79.8}}{--} &
\rc{57.0}{--} & \rc{65.1}{--} & \rc{75.6}{--} & \rc{85.3}{--} \\
\addlinespace[1pt]

\multicolumn{8}{@{}l}{\textbf{Ours}}\\
\textbf{PriorPose} & \cmark &
\rc{\textbf{84.1}}{\underline{94.3}} & \rc{\textbf{81.2}}{\textbf{92.5}} &
\rc{\textbf{61.8}}{\textbf{81.1}} & \rc{\textbf{68.2}}{\textbf{85.0}} &
\rc{\underline{81.3}}{\textbf{87.9}} & \rc{\textbf{88.9}}{\textbf{92.6}} \\
\bottomrule
\end{tabular}
\end{table}

\subsection{Comparison with State-of-the-Art Methods}
\label{sec:sota}

\begin{table}[t]
\caption{Performance comparison with state-of-the-art methods on HouseCat6D. Best is in \textbf{bold}, second-best is \underline{underlined}.}
\label{tab:housecat6d}
\centering
\scriptsize
\setlength{\tabcolsep}{2.2pt}
\renewcommand{\arraystretch}{1.08}

\begin{tabular}{@{}
  >{\raggedright\arraybackslash}p{2.85cm}
  c
  cccccc
@{}}
\toprule
\textbf{Method} & \textbf{Prior} &
\multicolumn{1}{c}{IoU$_{25}$} &
\multicolumn{1}{c}{IoU$_{50}$} &
\multicolumn{1}{c}{$5^{\circ}$2cm} &
\multicolumn{1}{c}{$5^{\circ}$5cm} &
\multicolumn{1}{c}{$10^{\circ}$2cm} &
\multicolumn{1}{c}{$10^{\circ}$5cm} \\
\midrule
VI-Net\cite{vinet} & \xmark & 80.7 & 56.4 & 8.4 & 10.3 & 20.5 & 29.1 \\
SecondPose\cite{secondpose} & \xmark & 83.7 & 66.1 & 11.0 & 13.4 & 25.3 & 35.7 \\
AG-Pose\cite{lin2024agpose} & \xmark & 81.8 & 62.5 & 11.5 & 12.0 & 32.7 & 35.8 \\
SpherePose\cite{ren2025spherepose} & \xmark & 88.8 & 72.2 & 19.3 & \underline{25.9} & 40.9 & 55.3 \\
SpotPose\cite{ren2025spotpose} & \xmark & \underline{89.1} & 77.0 & 23.8 & 24.5 & 52.3 & 54.8 \\
DPDN\cite{lin2022dpdn} & \cmark &
\multicolumn{1}{c}{--} & 56.2 & 6.4 & 6.9 & 22.2 & 25.8 \\
GCE-Pose\cite{li2025gcepose} & \cmark &
\multicolumn{1}{c}{--} & \underline{79.2} & \underline{24.8} & 25.7 & \underline{55.4} & \textbf{58.4} \\
\textbf{PriorPose} & \cmark &
\textbf{89.6} & \textbf{80.1} & \textbf{26.5} & \textbf{27.3} & \textbf{56.0} & \underline{58.0} \\
\bottomrule
\end{tabular}
\end{table}

\paragraph{REAL275 and CAMERA25.}
We use $5^\circ2$\,cm and $5^\circ5$\,cm as the main pose metrics, as they require accurate rotation and centimeter-level translation.
On REAL275, \textsc{PriorPose} achieves $61.8$ and $68.2$, improving over SpotPose~\cite{ren2025spotpose} by $+2.1$ and $+3.4$ points at these thresholds.
On CAMERA25, our method reaches $81.1$ and $85.0$ and remains competitive under more permissive criteria.
The consistent gains on both synthetic and real domains suggest more stable correspondences under domain shift; on REAL275, the improvements under strict thresholds further indicate robustness to sensor noise and outliers. We attribute this to keeping the prior explicit and jointly optimizing deformation and alignment, which mitigates error cascades in serial deform--then--align pipelines.

\paragraph{HouseCat6D.}
HouseCat6D features larger intra-class variation, frequent occlusions, and category symmetries.
As shown in Table~\ref{tab:housecat6d}, \textsc{PriorPose} achieves $26.5$ and $27.3$ mAP at $5^\circ2$\,cm and $5^\circ5$\,cm, improving over SpotPose~\cite{ren2025spotpose} by $+2.7$ and $+2.8$ points, and over GCE-Pose~\cite{li2025gcepose} by $+1.7$ and $+1.6$ points at the same thresholds.
The gains on strict pose mAP, together with improved IoU$_{25}$/IoU$_{50}$, are consistent with our design for partial and noisy depth: geometry-quality gating stabilizes feature fusion under outliers, while shape-consistency losses couple correspondence, deformation, and pose to reduce error propagation.

\subsection{Larger-Category Evaluation}
\label{sec:larger_category}

We further evaluate \textsc{PriorPose} on Omni6DPose~\cite{zhang2024omni6dpose} and PACE~\cite{you2024pace} to test whether the reference-guided joint formulation remains effective beyond the standard NOCS-style setting.
We follow the official category-level protocols and keep the core architecture unchanged.
Table~\ref{tab:large_category} reports representative metrics; full results are provided in the supplementary material.

\begin{table}[t]
\centering
\caption{Larger-category evaluation on Omni6DPose and PACE. We report representative official metrics for each benchmark.}
\label{tab:large_category}
\scriptsize
\setlength{\tabcolsep}{3.2pt}
\renewcommand{\arraystretch}{0.98}
\begin{tabular}{llcccc}
\toprule
Benchmark & Method & IoU$_{25}$ & IoU$_{50}$ & Strict & Relaxed \\ 
\midrule
Omni6DPose & AG-Pose~\cite{lin2024agpose} & 42.5 & 22.7 & 11.5 & 25.0 \\
Omni6DPose & GenPose++~\cite{zhang2024omni6dpose} & 39.0 & 19.1 & 15.1 & 29.4 \\
Omni6DPose & \textbf{PriorPose} & \textbf{46.3} & \textbf{26.4} & \textbf{16.4} & \textbf{30.8} \\
\midrule
PACE & CPPF++~\cite{you2024pace} & 44.5 & 4.4 & 9.9 & 24.9 \\
PACE & AG-Pose~\cite{lin2024agpose} & 69.4 & 28.1 & 19.5 & 39.7 \\
PACE & \textbf{PriorPose} & \textbf{75.1} & \textbf{35.5} & \textbf{21.8} & \textbf{43.3} \\
\bottomrule
\end{tabular}
\end{table}

For Omni6DPose, Strict/Relaxed denote VUS at $5^\circ$\,5cm/$10^\circ$\,5cm.
For PACE, Strict/Relaxed denote AP$_{R,t}$20/5 and AP$_{R,t}$60/15.
The results provide additional evidence under larger category diversity, while our conclusions remain limited to the evaluated RGB-D category-level protocols.

\subsection{Cross-dataset Generalization on Wild6D}
\label{sec:wild6d}

We evaluate zero-shot cross-dataset generalization on Wild6D~\cite{fu2022category}: all methods are trained on NOCS and tested on Wild6D without fine-tuning.
As shown in Table~\ref{tab:wild6d}, \textsc{PriorPose} improves over AG-Pose on both IoU and pose metrics, suggesting that explicit prior guidance helps stabilize correspondences under clutter and domain shift.

\begin{table}[t]
\centering
\caption{Zero-shot evaluation on Wild6D. All methods are trained on NOCS only and evaluated on Wild6D without fine-tuning.}
\label{tab:wild6d}
\scriptsize
\setlength{\tabcolsep}{1.8pt}
\renewcommand{\arraystretch}{0.98}
\begin{tabular}{@{}l c c c c c c c@{}}
\toprule
\textbf{Method} & \textbf{Prior} &
IoU$_{50}$ & IoU$_{75}$ &
$5^\circ2$cm & $5^\circ5$cm &
$10^\circ2$cm & $10^\circ5$cm \\
\midrule
AG-Pose~\cite{lin2024agpose}   & \xmark & 86.0 & 57.0 & 41.8 & 44.9 &  49.0 & 54.2 \\ 
\textbf{PriorPose}            & \cmark & \textbf{87.2} & \textbf{62.4}	& \textbf{43.1}	& \textbf{46.3}	 & \textbf{52.6}	& \textbf{56.2}\\
\bottomrule
\end{tabular}
\end{table}

\subsection{Ablation Studies}
\label{sec:ablation}
\noindent\textbf{Protocol.}
All ablations are on REAL275 with the standard split and the same evaluation metrics as Table~\ref{tab:main_results}.
Unless specified, we keep data, schedule, and heads fixed, and only toggle the prior branch, loss terms, or fusion settings.



\noindent\textbf{A. Joint formulation and prior complexity.}
We isolate whether the gain comes from simply adding a prior or from jointly coupling deformation and alignment.
\emph{No-Prior} removes the prior branch.
\emph{Serial D$\rightarrow$A} uses the same prior and seeded transformer, but first predicts a deformed prior and then treats it as a fixed reference for NOCS and pose estimation.
\emph{Gauss. Prior} replaces the structured mean-shape prior with a fixed category-wise full-covariance Gaussian prior.
\emph{Joint w/o SC} removes the shape-consistency losses from the joint model.

\begin{table}[t]
\centering
\caption{Controlled ablations on REAL275. SC denotes shape consistency.}
\label{tab:controlled_ablation}
\scriptsize
\setlength{\tabcolsep}{3.0pt}
\renewcommand{\arraystretch}{0.98}
\begin{tabular}{lcccc}
\toprule
Variant & $5^\circ$\,2cm & $5^\circ$\,5cm & $10^\circ$\,2cm & $10^\circ$\,5cm  \\
\midrule
No-Prior & 57.2 & 64.4 & 76.5 & 85.7 \\
Serial D$\rightarrow$A & 58.1 & 64.5 & 78.2 & 86.7 \\
Gauss. Prior & 59.9 & 65.3 & 78.4 & 86.4 \\
Joint w/o SC & 60.2 & 65.5 & 80.7 & 87.3 \\
Full & \textbf{61.8} & \textbf{68.2} & \textbf{81.3} & \textbf{88.9} \\
\bottomrule
\end{tabular}
\end{table}

\emph{Observation.}
Serial D$\rightarrow$A improves over No-Prior but remains below Joint w/o SC, showing that the gain is not from the prior alone.
Gauss. Prior also improves over No-Prior but underperforms the structured prior, suggesting that surface topology and part layout provide useful correspondence guidance.
Full further improves over Joint w/o SC, confirming the benefit of shape consistency.

\noindent\textbf{B. Effect of the shape-consistency losses.}
We ablate the three shape-related losses that couple deformation, correspondence, and pose.
All variants keep the same NOCS and pose supervision; we only change the reconstruction and consistency terms.
\emph{w/o Recon Loss} disables the deformation branch and removes all shape-related losses.
\emph{w/o Shape Consistency} enables the deformation head and uses only the reconstruction loss $\mathcal{L}_{\mathrm{rec}}$ on the canonical instance.
\emph{Canonical Consistency} additionally applies the canonical-space loss $\mathcal{L}_{\mathrm{can}}$ to tie the NOCS field to the reconstructed canonical instance.
\emph{Camera Consistency} instead uses the camera-space loss $\mathcal{L}_{\mathrm{cam}}$ to enforce agreement between the pose-placed reconstruction and the observed geometry.
\emph{Full} combines $\mathcal{L}_{\mathrm{rec}}$, $\mathcal{L}_{\mathrm{can}}$, and $\mathcal{L}_{\mathrm{cam}}$.

\begin{table}[t]
\centering
\scriptsize
\caption{Ablation studies on the reconstruction loss and shape consistency losses.}
\setlength{\tabcolsep}{1.5pt}
\begin{tabular}{lccccccc}
\toprule
Variant &
$\mathcal{L}_{\mathrm{rec}}$ &
$\mathcal{L}_{\mathrm{can}}$ &
$\mathcal{L}_{\mathrm{cam}}$ &
$5^\circ2$cm &
$5^\circ5$cm &
$10^\circ2$cm & $10^\circ5$cm \\
\midrule
w/o Recon Loss          & \xmark & \xmark & \xmark & 58.1 & 64.7 & 78.1 & 86.2 \\
w/o Shape Consistency   & \cmark & \xmark & \xmark & 60.2 & 65.5 & 80.7 & 87.3 \\
+ Canonical Consistency & \cmark & \cmark & \xmark & 61.2 & 65.9 & 81.2 & 87.6 \\
+ Camera Consistency    & \cmark & \xmark & \cmark & 61.4 & 67.5 & 79.8 & 87.8 \\
Full (ours)             & \cmark & \cmark & \cmark & \textbf{61.8} & \textbf{68.2} & \textbf{81.3} & \textbf{88.9} \\
\bottomrule
\end{tabular}
\label{tab:ablation_losses}
\end{table}

\emph{Observation.}
Without reconstruction and consistency (w/o Recon Loss), the model yields the lowest pose accuracy, indicating limited canonical regularization from NOCS and pose supervision alone.
Adding reconstruction only (w/o Shape Consistency) improves all thresholds by $+2.1/+0.8/+2.6/+1.1$ at $5^\circ2$\,cm, $5^\circ5$\,cm, $10^\circ2$\,cm, and $10^\circ5$\,cm.
Canonical consistency improves most metrics, while camera-space consistency mainly improves $5^\circ5$\,cm and $10^\circ5$\,cm but slightly decreases $10^\circ2$\,cm. Combining both yields the best overall performance.
In particular, Full improves over w/o Shape Consistency by $+1.6/+2.7/+0.6/+1.6$, supporting the complementarity of canonical-space and camera-space constraints for coupling deformation, correspondence, and pose.

\noindent\textbf{C. Fusion type, neighborhood size, and geometry-quality bias.}
Finally, we compare dense versus seeded fusion and examine the sensitivity to the kNN size $k$ and the score-bias weight $\gamma$ used in seeded attention.
\emph{Dense self-attn} replaces the seeded transformer with a standard dense self-attention block over the concatenated observation and prior tokens.
\emph{Seeded} variants use geometry-aware seeds as in Sec.~\ref{sec:fusion}, with different choices of $k$ and $\gamma$; when $\gamma{=}0$, the geometry-quality bias is disabled.

\begin{table}[t]
\centering
\footnotesize
\caption{Ablation studies on fusion type and neighborhood configuration.}
\setlength{\tabcolsep}{1.5pt}
\begin{tabular}{lcccccccc}
\toprule
Variant & Fusion & $k$ & $\gamma$ &
$5^\circ2$cm & $5^\circ5$cm & $10^\circ2$cm & $10^\circ5$cm \\
\midrule
Dense self-attn              & dense  & -- & --   & 60.4 & 66.5 & 79.7 & 87.5 \\
Seeded (no score)            & seeded & 32 & 0.0  & 61.2 & 66.4 & 79.4 & 87.2 \\
Seeded ($k{=}16$)            & seeded & 16 & 0.2  & 61.0 & 66.6 & 79.5 & 86.2 \\
Seeded ($k{=}32$)   & seeded & 32 & 0.2  & \textbf{61.8} & \textbf{68.2} & \textbf{81.3} & \textbf{88.9} \\
Seeded ($k{=}64$)            & seeded & 64 & 0.2  & 61.2 & 67.1 & 80.2 & 88.3 \\
\bottomrule
\end{tabular}
\label{tab:ablation_fusion}
\end{table}

\emph{Observation.}
Dense self-attention already gives strong performance, but seeded fusion achieves better accuracy with a compact seed pool.
With the default configuration ($k{=}32,\gamma{=}0.2$), seeded fusion improves over dense self-attention by $+1.4/+1.7/+1.6/+1.4$ at $5^\circ2$\,cm, $5^\circ5$\,cm, $10^\circ2$\,cm, and $10^\circ5$\,cm.
Disabling the geometry-quality bias reduces performance across most thresholds, indicating that geometry-quality routing helps suppress unreliable points in sparse depth observations.
The neighborhood size also matters: both $k{=}16$ and $k{=}64$ underperform $k{=}32$, suggesting that moderate local neighborhoods provide the best trade-off between locality and cross-stream information flow.

\subsection{Qualitative and Failure Analysis}

\begin{figure}[htbp]
  \centering
  \includegraphics[width=1\linewidth]{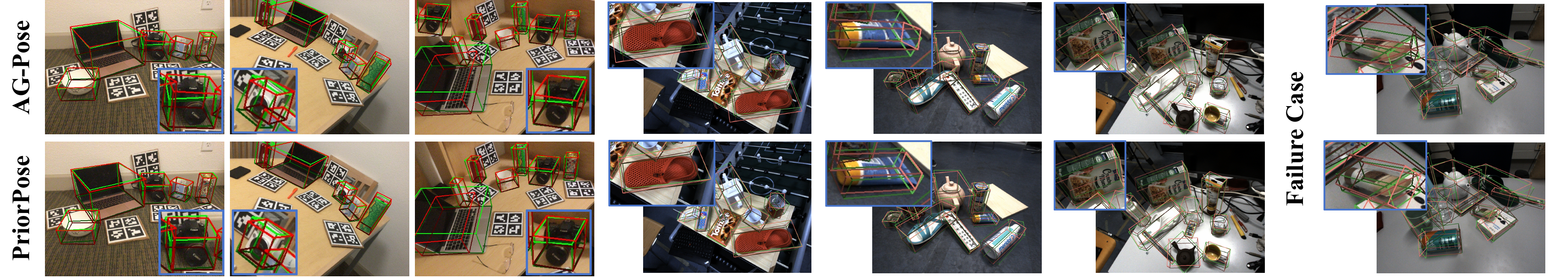}
  \caption{
Qualitative comparison with AG-Pose on REAL275 and HouseCat6D (top: AG-Pose, bottom: \textsc{PriorPose}).
Red and green boxes denote predicted and ground-truth poses, respectively.
Blue insets highlight challenging regions.
Right: a failure case on HouseCat6D, where both methods struggle under heavy clutter and occlusion.
}
  \label{fig:visual2}
\end{figure}

\textbf{Visualization.}
We visualize qualitative results on REAL275 and HouseCat6D.
Figure~\ref{fig:visual2} compares \textsc{PriorPose} with AG-Pose~\cite{lin2024agpose} and includes a failure case under heavy clutter and occlusion. Additional visualizations of reconstructed canonical instances are provided in the supplementary material.

\section{Conclusion}
In this paper, we present \textsc{PriorPose}, a reference-guided correspondence framework for category-level object pose estimation. By introducing an explicit category prior and a seeded transformer, our method fuses prior and partial observations in a shared feature space and jointly predicts per-point NOCS, canonical deformation, and pose. A shape-consistent training objective in canonical and camera space further couples deformation and alignment, leading to more stable correspondences under occlusion, symmetries, and large intra-class variation. Extensive experiments and ablations demonstrate the effectiveness of our design.

\section{Acknowledgements}
This work was supported by the Open Fund of National Key Laboratory of Deep Space Exploration (Grant NKDSEL2025008).
\bibliographystyle{splncs04}
\bibliography{main}
\end{document}